\documentclass[conference]{IEEEtran}
\usepackage{cite}
\usepackage{amsmath,amssymb,amsfonts}
\usepackage{algorithmic}
\usepackage{graphicx}
\usepackage{textcomp}
\usepackage{xcolor}
\usepackage[utf8]{inputenc}
\usepackage{hyperref}
\usepackage{url}
\usepackage{booktabs}
\usepackage{multirow}
\usepackage{caption,subcaption}
\def\BibTeX{{\rm B\kern-.05em{\sc i\kern-.025em b}\kern-.08em
    T\kern-.1667em\lower.7ex\hbox{E}\kern-.125emX}}
\begin{document}

\title{Large Scale Language Modeling: Converging on 40GB of Text in Four Hours}

\author{
\IEEEauthorblockN{Raul Puri}
\IEEEauthorblockA{
{NVIDIA}\\
raulp@nvidia.com}
\and
\IEEEauthorblockN{Robert Kirby}
\IEEEauthorblockA{
{NVIDIA}\\
rkirby@nvidia.com}
\and
\IEEEauthorblockN{Nikolai Yakovenko}
\IEEEauthorblockA{
{NVIDIA}\\
nyakovenko@nvidia.com}
\and
\IEEEauthorblockN{Bryan Catanzaro}
\IEEEauthorblockA{
{NVIDIA}\\
bcatanzaro@nvidia.com}
}

\maketitle

\begin{abstract}

Recent work has shown how to train Convolutional Neural Networks (CNNs) rapidly on large image datasets\cite{Akiba2017}, then transfer the knowledge gained from these models to a variety of tasks\cite{ImageNetTransfer2018}. Following~\cite{Radford2017}, in this work, we demonstrate similar scalability and transfer for Recurrent Neural Networks (RNNs) for Natural Language tasks. 

By utilizing mixed precision arithmetic and a 32k batch size  distributed across 128 NVIDIA Tesla V100 GPUs, we are able to train a character-level 4096-dimension multiplicative LSTM (mLSTM) \cite{Krause2016} for unsupervised text reconstruction over 3 epochs of the 40 GB Amazon Reviews dataset \cite{McAuley2015} in four hours. 
This runtime compares favorably with previous work taking one month to train the same size and configuration for one epoch over the same dataset\cite{Radford2017}. 

Converging large batch RNN models can be challenging. Recent work has suggested scaling the learning rate as a function of batch size, but we find that simply scaling the learning rate as a function of batch size leads either to significantly worse convergence or immediate divergence for this problem. 
We provide a learning rate schedule that allows our model to converge with a 32k batch size. 

% [cost and accessibility]
Since our model converges over the Amazon Reviews dataset in hours, and our compute requirement of 128 Tesla V100 GPUs, while substantial, is commercially available, this work opens up large scale unsupervised NLP training to most commercial applications and deep learning researchers\footnote{Our code is publicly available: https://github.com/NVIDIA/sentiment-discovery}. A model can be trained over most public or private text datasets overnight. 

\end{abstract}

\section{Introduction}
In recent years, deep learning has been successfully applied to many problems. The successful use of transfer learning for computer vision problems has enabled many applications: large CNNs such as VGG \cite{Simonyan2014} and ResNets \cite{He2015} are pre-trained on a large image dataset such as ImageNet \cite{Deng2009,Russakovsky2014} and then utilized as the backbone for other computer vision tasks. These models are able to extract meaningful features for new tasks without needing to be trained from scratch for each task \cite{DeCAF,CNNTransfer,MaskRCNN,ImageNetTransfer2018}.

Recent work has shown promising results from unsupervised language modeling, followed by transfer learning to natural language tasks~\cite{Radford2017}, \cite{Radford2018}.
However, neural language models have not benefited from scale and transfer learning in the same way as convolutional image models. Historically, natural language leverages large scale transfer learning through the use of word embedding pretraining on large corpora \cite{Turian2010,Mikolov2013,Pennington2014}. Transferring only the embeddings limits the scope of the transfer, since word embeddings do not capture sequential information in a section of text. We would like to transfer whole NLP models capable of processing a text sequence.

However, transfer learning in this context is difficult because of the time it takes to train large language models on large datasets. Several recent publications seek to address long training times by leveraging distributed data parallelism and increasing the effective batch size during training \cite{Goyal2017,Akiba2017,You2017a,You2017b,Ott2018}, taking advantage of advances in distributed deep learning and improvements in the memory size and compute capability of available high performance computing (HPC) resources. This work often focuses on computer vision and rarely on natural language tasks, let alone RNN-based language models, which are numerically difficult to train and suffer from poor parallelization due to their sequential nature. We do have evidence that RNNs for language modeling, speech recognition, and neural machine translation continue to provide accuracy improvements as they are trained on larger datasets~\cite{Hestness2017}. Accordingly, techniques for efficiently training large RNN models will lead to improved accuracy on many natural language tasks.

We focus on training a single-layer 4096 neuron multiplicative LSTM-based character language model \cite{Krause2016} on the Amazon Reviews dataset, one of the largest publicly-available NLP datasets, and transfer the model to the downstream tasks of sentiment classification on the Binary Stanford Sentiment Treebank (SST) and IMDB movie review datasets. We train our recurrent models with mixed precision FP16/FP32 arithmetic, which speeds up training on a single V100 by 4.2X over training in FP32. 

We then train the mixed precision model using a 32k batch size via distributed data parallelism across 128 GPUs. This achieves a 109x increase in training data throughput relative to the single GPU case. However, with such a large batch size, we require additional epochs to train the model to a similar level of accuracy, bringing the total training time to 4 hours. 

In addition, we train a 8192 neuron mLSTM capable of beating state of the art performance in Amazon review language modeling with a bits per character (BPC) of 1.038 and SST classification accuracy of 93.8\%. 

We analyze how distributed data parallelism scales with larger models. While utilizing distributed data parallelism for training RNNs, we observe some problems common to training with large batches. We investigate the relationship between dataset size, batch size, and learning rate schedule to investigate how to effectively use large batch training to train models on commonly available large NLP datasets. 

\section{Language Model Pretraining and Transfer}
Separately trained word embeddings\cite{Turian2010,Mikolov2013,Pennington2014} are commonly used to transfer learning from large datasets to specific tasks. However, word embeddings function only as a lookup table for in-vocabulary words. They do not transfer well to multi-word sequences and contexts. 

Works such as Semi-supervised Sequence Learning \cite{Dai2015}, context2vec \cite{Melamud2016}, Contextualized Word Vectors (CoVe) \cite{McCann2017}, and Deep Contextualized Word Representations (ELMo) \cite{Peters2018} seek to remedy this by computing embeddings of words in a sequence using a pretrained recurrent neural language model. In these approaches, the surrounding words provide context which is used to produce an embedding that represents the meaning of a given word. These works approach the transfer learning problem with a whole neural language model capable of modeling the compositional nature of language rather than a lookup table that considers all words independently.

%Recent works such as context2vec \cite{Melamud2016}, Contextualized Word Vectors (CoVe) \cite{McCann2017}, and Deep Contextualized Word Representations (ELMo) \cite{Peters2018} seek to remedy this by computing embeddings of words in a sequence using a pretrained recurrent neural language model. In these approaches, the surrounding words provide context which is used to produce an embedding that represents the meaning of a given word. These works approach the transfer learning problem with a whole neural language model capable of modeling the compositional nature of language rather than a lookup table that considers all words independently.

This pretraining and transfer work has motivated follow on works trying to increase the scope of neural language model pretraining and transfer \cite{Radford2017,Howard2018,Subramanian2018,Cer2018,Liu2018,Radford2018}, in which the authors explore new types of language models, multiple types of language model pretraining, and the effect these two have on a wide variety of down stream language tasks. A common theme between these different research efforts, however, is that downstream transfer success is predicated on the pretraining corpus size. Larger text corpora produce more powerful language models, which then improve transfer learning.

\subsection{Pretraining Tasks and Datasets}

As part of pretraining there are three components that determine pretraining success: the task used for pretraining, pretraining dataset quality, and pretraining dataset size.

The former requires careful consideration as it affects the other two. A number of language pretraining tasks can be considered generative pretraining tasks, where the language models are trained to generate some language  as output. Some of these include sequence to sequence (Seq2Seq) tasks such as Skip-Thought pretraining \cite{Kiros2015,Subramanian2018} and Neural Machine Translation \cite{Ramachandran2016,Subramanian2018}. However, we instead choose to focus on unsupervised text reconstruction as our pretraining task: predict the next character of text, given the previous characters. Text reconstruction captures the fundamental components of sequence modeling required by other language modeling tasks.

With text reconstruction, the data provides its own labels, and given the data has undergone reasonable cleaning, we can focus on dataset size rather than dataset type or quality. Several corpora successfully utilized for unsupervised text pretraining in prior work are the BooksCorpus \cite{Zhu2015}, GigaWord \cite{Parker2011}, 1-Billion Word \cite{Chelba2013}, and Amazon Reviews \cite{McAuley2015} datasets. Similar to \cite{Radford2017,Gray2017}, we focus our pretraining efforts on the largest of the four datasets (see Fig. \ref{tab:dataset_size}), by training a mLSTM on an aggressively deduplicated copy of the Amazon Reviews dataset totaling 82 million reviews (40GB). The generality of our task and the size of our dataset allow the insights developed in this work to be applied to other large scale language tasks.
\begin{figure}[t!] 
  \begin{center}
    \begin{tabular}{l|r}
      \toprule % <-- Toprule here
      \textbf{Dataset} & \textbf{corpus size (GBs)}\\
      \midrule % <-- Midrule here
      1-Billion Word & 3\\
      BooksCorpus & 5\\
      GigaWord & 26\\
      Amazon Reviews Dataset & 41\\
      \bottomrule % <-- Bottomrule here
    \end{tabular}
    \caption{Various large language corpora and their size}
    \label{tab:dataset_size}
  \end{center}
\end{figure}

\section{Large Batch Training} \label{sec:large_methods}

Given the size of the Amazon corpus, pretraining a large state of the art neural language model is a time consuming process. Running such a workload on a single GPU is not practical, as state of the art models tend to be large and can only fit a modest training batch size per GPU. In order to enable effective pretraining and transfer of large language models, we employ multi-GPU parallelism. We focus on scaling to multiple GPUs with data parallelism, meaning that we partition the batch during training across multiple GPUs. We don't use model parallelism, which partitions the neural network itself across multiple processors, because it's less flexible and places more constraints on software, although it remains an interesting avenue for further parallelism.

We use synchronous data parallelism, where a large batch is distributed evenly amongst all participating worker processes, at which point the worker processes run forward and backward propagation, communicate the resulting gradients with each other, and update the model before receiving a new data batch. Depending on model size and communication latency, data parallelism allows for near linear speed up by scaling batch size linearly with respect to the number of available GPUs. Taking advantage of such scaling, the Computer Vision community has been able to reduce the training time of AlexNet and ResNet-50 models on the ImageNet benchmark from hours to the order of minutes \cite{Goyal2017,Akiba2017, You2017a,You2017b}. 

However, these projects have focused on convolutional networks for image classification, and comparatively less work has been published on large batch training of language models. Ott et. al \cite{Ott2018} employ data parallelism to speed up Seq2Seq neural machine translation. However, similar to prior work, Ott et. al train convolutional models with large batches. 

In order to enable large batch pretraining of an arbitrary language model it is important to explicitly analyze the effects of large batch training with RNN-based language models. The sequential nature of recurrent neural networks makes the training landscape difficult to optimize, due to saddle points, local minima, and numerical instabilities in the RNN computation itself \cite{LeCun2012,Pascanu2012,Karpathy2015}. These complexities necessitate analysis of large batch training with RNNs.

Large batch training is itself not without difficulties. Identical hyperparameters at different batch sizes regularly produce models which generalize differently. Recent work analyzes the relationship between large batch size, learning rates, and generalization, showing how to achieve similar evaluation results when training across different batch sizes \cite{Smith2017a,Smith2017b,Goyal2017}. 

By analyzing the noise scale of gradient-descent optimization, these methods modify learning rate $\epsilon$ proportionally to batch size $B$, with a linear scaling rule $\epsilon \propto B$ provided that $B \ll N$, where $N$ is the dataset size. The authors find that learning rate scaling leads to models that generalize well across various batch sizes. Additionally, Smith et. al \cite{Smith2017a,Smith2017b} proposed scaling momentum as a function of batch size; however, we do not investigate such scaling in this work. 

In order to enable large batch training of RNN language models, we explore the effects of this linear scaling rule as well as a softer square root scaling rule $\epsilon \propto \sqrt{B}$ proposed by Hoffer et. al \cite{Hoffer2017}.

Additionally, we investigate the scalability of data parallelism with different interconnects and model sizes, so as to assess the effectiveness of data parallelism for an arbitrary neural language model.

\section{Distributed Deep Learning Setup} \label{DDL_setup}

We use NVIDIA DGX1-V systems built from 16 GB Tesla V100 GPUs. For intra-node and inter-node communication we leverage the {NCCL}2 (NVIDIA Collective Communications) library which uses the DGX1-V's underlying NVLink and InfiniBand connections for GPU to GPU communication. 

We do not use a central parameter server for managing gradient reduction and updating the model. In order to efficiently perform updates to the model, the group of worker processes perform a ring reduce of the gradients, and each worker independently updates the model parameters. Crucial to reducing the necessary communication bandwidth, the library also supports communication of FP16 values natively with no FP16 emulation overhead when reducing FP16 parameter gradients across GPUs.

\section{Mixed Precision Training} \label{ssec:fp16_methods}

FP16 is not only useful for reducing communication overhead, it also plays a key role in directly accelerating training on processors like the V100 that support higher throughput mixed-precision arithmetic. The V100 provides 15.6 TFlops in single precision, but 125 TFlops with mixed-precision arithmetic (FP16 storage and multiplication, FP32 accumulation).
Using FP16 reduces the dynamic range and precision of the computations being performed. This presents a unique set of training difficulties, which, if not addressed, lead to convergence issues while training. 

Drawing from \cite{Micikevicius2017,AutoLossScale}, we use automatic loss scaling to effectively increase the dynamic range of the training process. Automatic loss scaling seeks to ameliorate numerical underflow by multiplying the training loss by a scalar "loss scale" factor $\alpha>1$, performing backpropagation with all intermediary gradients multiplied by $\alpha$, and dividing the final weight gradients by $\alpha$. This multiplication shifts small gradient values into the range permitted by FP16, thereby ensuring that they do not vanish during back propagation. 

We choose $\alpha$ dynamically by starting at a large value, performing backpropagation, and checking for an overflow in the weight gradients. If an overflow is detected, then the weight update for the batch is skipped, and $\alpha$ is halved. After the algorithm finds a suitable $\alpha$, it tries to increase $\alpha$ after a sufficient number of iterations have passed without overflow, and again backs off if overflow occurs. The algorithm repeats this process throughout training, iteratively updating the loss scale, hence the name automatic loss scaling.

Without automatic loss scaling, we found that our models did not train to convergence. Although the computationally intensive parts of training were performed in mixed precision, a minority of the work still remained in FP32 in order to converge properly:
\begin{itemize}
\item Gradients are accumulated into a “master” FP32 copy of the parameters. The division by $\alpha$ occurs on the gradients of these master copies.
\item Reductions are performed in FP32; it only takes a few large values to cause an overflow in FP16.
\item Accumulation of the summation in the $\ell$2 norm computation required by weight normalization should be done in FP32 to avoid overflow. The final norm value is output in FP16.
\item Softmax loss is computed in FP32, operating on FP32 logits in order to avoid numerical issues when exponentiating FP16 values.
\end{itemize}

These techniques working in conjunction allowed for successful training of the mLSTM language model in mixed precision.

\section{Experiments}

All experiments are set up following \cite{Radford2017} and run with Pytorch's v0.4 release \cite{PyTorch}. The Amazon Reviews dataset is shuffled and split into training, validation, and test sets. The model is trained using truncated backpropagation through time (TBTT) \cite{Sutskever2013} on sequences of 256 characters.  We persist hidden state across each minibatch during training and evaluation.  

\subsection{Data Sharding}

In order to create the training, validation, and test sets, the dataset is split proportionally by a ratio of 1000, 1, and 1 allocated for train, validation, and test sets respectively. Within these sets we create batch size $B$ number of shards for evaluation, and $max(1000,B)$ shards for training. A shard is defined as a subset of strings sampled without replacement from one of the dataset splits; this subset is concatenated together into one large string to form a shard. These shards are used for all training epochs with no further shuffling. Hidden state is initialized to zero at the beginning of a shard and persisted throughout the shard.

When constructing a minibatch $D_{i_j}$, data is sampled such that between two consecutive minibatches $i$ and $i+1$, minibatch index $j$ contains contiguous subsequences from within a shard. This contiguity across minibatches enables hidden state persistence across truncated subsequences in TBTT.

\subsection{Weight Normalization}
In order to aid with convergence during training time, we applied weight normalization \cite{Salimans2016} to the LSTM parameters only, following \cite{Radford2017}. This includes the 4 hidden$\rightarrow$hidden and input$\rightarrow$hidden parameters of the multiplicative LSTM. Weight normalization was not applied to the bias terms.

\subsection{Optimization and Learning Rate (LR) schedule}
As in \cite{Radford2017}, Adam \cite{Kingma2014} is utilized for optimization, along with a learning rate schedule that decays linearly to zero over the course of training. For a global batch size of 128 a learning rate of 5e-4 is used, and is scaled up according to the batch size using either the linear or square root scaling rule.

\subsection{Evaluation}
Two metrics for evaluating training are considered: 
\begin{enumerate}
    \item A bits per character (BPC) metric calculated on the immediate task of predicting the next character given the current character on the Amazon Reviews test set. We calculate the average BPC across 16 random shards of the test set by using an evaluation batch size $B$ of 16. Since our model operates directly on character-level tokens, calculation of BPC is simply $l\cdotp \log_2e$ where $l$ is the softmax cross entropy loss averaged over the entire sequence.
    \item Accuracy from the downstream tasks of binary sentiment classification on the Binary SST, and IMDB Movie Review datasets. To perform transfer the model weights are taken at the end of Amazon training, frozen, and used to featurize text samples from the classification dataset. A simple binary logistic regression classifier from scikit-learn \cite{scikit-learn} is trained to classify these text features as having positive or negative sentiment. The transfer process is negligible computationally because of the simple model we use on the downstream task.
\end{enumerate}

\section{Analysis of Mixed Precision vs FP32 training}\label{sec:f16_analysis}

\begin{figure}[t!] 
  \captionsetup[subfigure]{justification=raggedright,singlelinecheck=false}
  \centering
  \begin{subfigure}[b]{\columnwidth}
    \centering
    \caption{\label{fp16_fp32:graph}}
    \includegraphics[width=\columnwidth]{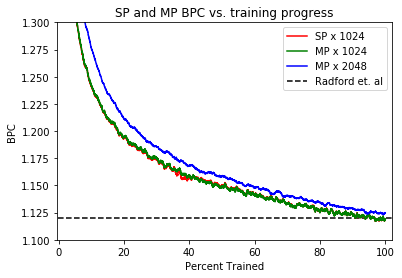}
  \end{subfigure}
  \begin{subfigure}[b]{\columnwidth}
    \caption{\label{fp16_fp32:table}}
    \begin{tabular}{l|r|r|r|r|r|r}
      \toprule % <-- Toprule here
      \textbf{Type} & \textbf{Batch} & \textbf{LR} & \textbf{Time} & \textbf{BPC} & \textbf{SST} & \textbf{IMDB}\\
      \midrule % <-- Midrule here
      \textbf{SP} & \textbf{128} & \textbf{5e-4} & \textbf{1 month} & \textbf{1.12} & \textbf{91.9} & \textbf{92.8} \\
      SP & 1024 & 1.2e-3 & 73.8 hr & 1.104 & 90.8 & 92.5\\
      MP & 1024 & 1.2e-3 & 24.2 hr & 1.108 & 91.5 & 91.7\\
      MP & 2048 & 2e-3 & 17.4 hr & 1.117 & 90.2 & 91.9 \\
      \bottomrule % <-- Bottomrule here
    \end{tabular}
    \end{subfigure}
    \caption{\subref{fp16_fp32:graph}) Training curves for mixed precision (MP) and single precision (SP) training \subref{fp16_fp32:table}) Test set evaluation comparison of single precision vs mixed precision training w.r.t. the Amazon BPC and binary sentiment classification accuracy baselines set by Radford et. al \cite{Radford2017}}
    \label{tab:fp16_fp32}
\end{figure}
Mixed precision training allows for faster computation as well as a 2x increase in effective batch size during training, because FP16 storage is 2x smaller. In this section we analyze performance gains and convergence for training networks with mixed precision arithmetic, comparing it to single precision training. This allows us to validate the correctness of the remaining experiments, which are all trained in mixed precision.

Using the techniques described in section \ref{DDL_setup} \& \ref{ssec:fp16_methods}, we train a model on the Amazon Reviews dataset using a full DGX1-V node with 8 GPUs. We initially begin with a batch size of 128 per GPU, for a global batch size of 1024, and compare the relative speedup granted by mixed precision arithmetic. Next, we quantify the benefits of the reduced memory footprint by doubling the batch size to 256 per GPU (2048 global) in order to better saturate the GPU. Additionally, we utilize the softer square root scaling rule \cite{Hoffer2017} to modify the learning rate as a function of batch size.

Figure \ref{tab:fp16_fp32} shows that training in mixed precision and single precision both produce similar training curves and converge to similar numbers for both language modeling and transfer evaluation. We find that moving to mixed precision not only achieves similar training results, but it also provides a 3x speedup in training. By taking advantage of the reduced memory footprint of FP16 storage, we increase the batch size two-fold to 256 per GPU, better saturating the GPU, and achieve an additional speedup of 40\% on top of our original speedup. This provides approximately a 4.2x speedup when switching from single precision arithmetic to mixed precision.

Overall, this yields a speed up from one month of training as in \cite{Radford2017} to 18 hours. We have accomplished this using 8 Tesla V100 GPUs, larger batch size, and mixed precision arithmetic.

\section{Distributed Data Parallel Scaling}\label{sec:DDP_scale}
To train a language model in hours, not in days, we further parallelize the training process by using multiple nodes and additional data parallelism. We first analyze the effect of communication overhead on the scalability of multi-GPU training at various batch sizes and processor counts.

\begin{figure}[h!]
  \captionsetup[subfigure]{justification=raggedright,singlelinecheck=false}
  \centering
  \begin{subfigure}[b]{\columnwidth}
  \caption{\label{scale:graph}}
  \centering\includegraphics[width=\columnwidth]{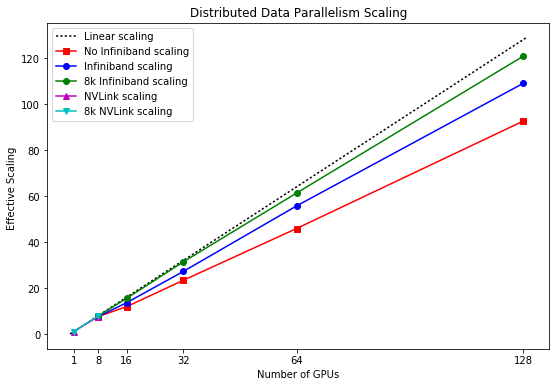}
  
  \end{subfigure}
  \begin{subfigure}[b]{\columnwidth}
  \caption{\label{scale:table}}
  \centering
    \begin{tabular}{l|r|r|r|r|r|r}
      \toprule % <-- Toprule here
      \multirow{2}{*}{\textbf{GPUs}} & \multicolumn{2}{c|}{w/o I.band} & \multicolumn{2}{c|}{w/ I.band} & \multicolumn{2}{c}{8192-d + I.band}\\
      \cline{2-7}
      & \textbf{s/iter} & \textbf{speed} & \textbf{s/iter} & \textbf{speed} & \textbf{s/iter} & \textbf{speed}\\
      \midrule % <-- Midrule here
      1 & .81 & 1x & .81 & 1x & 2.01 & 1x\\
      8 & .85 & 7.6x & .85 & 7.6x & 2.02 & 7.9x\\
      16 & 1.09 & 14.3x & .91 & 13.6x & 2.08 & 15.5x\\
      32 & 1.11 & 23.4x & .91 & 27.2x & 2.05 & 31.4x\\
      64 & 1.13 & 55.7x & .93 & 55.7x & 2.10 & 61.3x\\
      128 & 1.12 & 92.6x & .91 & 109x & 2.13 & 120.8x\\
      \bottomrule % <-- Bottomrule here
    \end{tabular}
    
    \end{subfigure}
    
    \caption{\subref{scale:graph}) Training time for 1 epoch of Amazon Reviews exhibits linear scaling relative to the single GPU case. \subref{scale:table}) Average per iteration times and relative speedup for distributed data parallel training with (and without) Infiniband.}
    
    \label{tab:gpu_scale}
    
\end{figure}

The model is trained in mixed precision on 1, 8, 16, 32, 64, and 128 GPUs with a local batch size of 256 batches/GPU and 8 GPUs/DGX1-V node. In Fig. \ref{tab:gpu_scale} we observe that NCCL2 provides near linear scaling with minimal overhead when scaling from 1 to 8 GPUs within a node. Infiniband efficiently handles inter-node communication for the 4096 neuron mLSTM with effectively constant overhead with respect to the number of participating nodes. This allows for a total speedup of 109x when scaling to 128 GPUs across 16 DGX1-V Nodes. More concretely, we complete one epoch of training on the Amazon reviews dataset in only 1.2 hours.

\subsection{Scaling Large Model Training}

\begin{figure}[t!] 
  \begin{center}
    \includegraphics[width=\columnwidth]{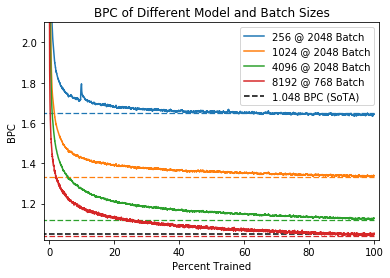}
    \caption{Training progress over one epoch of Amazon Reviews for mLSTM models at a particular dimension and batch size. Dashed lines indicate the evaluation BPC after one epoch of training, with State Of The Art (SOTA) evaluation results set by Gray et. al \cite{Gray2017}.}
    \label{fig:large_run}
  \end{center}
\end{figure}

Not every problem calls for training a 4096-d mLSTM. Smaller models will train faster and may converge to a good enough BPC, while larger models may be necessary for state of the art performance. To illustrate this, we train an mLSTM with hidden state sizes of 256, 1024, 4096, and 8192 and a global training batch size of 2048 split across 1 DGX1-V node and learning rate of 2e-3. In the case of the 8192-d hidden state mLSTM we use a per GPU batch size of 96 (768 total) due to memory constraints. In this experiment, we use a learning rate of 7.8e-4 that observes the square root scaling rule. In Fig. \ref{fig:large_run} we can see the benefit of training larger models, with the 8192-d mLSTM achieving state of the art language modeling comparable to \cite{Gray2017}, albeit at the cost of additional compute and memory.

We investigate the scalability of a larger 8192-d  mLSTM model compared to the baseline 4096-d mLSTM model in Fig. \ref{tab:gpu_scale}. The 8192-d model has 0.72 GB of parameters in FP16, while the 4096-d model has 0.18 GB of parameters. While training the 8192-d model on 128 GPUs, we see a speedup factor of 120.8x across 128 GPUs. Even though the larger model has correspondingly larger gradients, it is also more computationally intensive, leading to better scaling than the baseline model on the same hardware.

\section{Analysis of Large Batch training}

Distributed data parallel training allows for near linear scalability with respect to available GPUs by increasing the batch size. However, as seen already in this work (see section \ref{sec:f16_analysis}), training with large batches may run faster than training with small batches, but it may not converge to the same validation accuracy. Using the same setup as section \ref{sec:DDP_scale} and 8, 16, 32, 64, and 128 GPUs we take a look at how the learning rate schedule affects convergence.

\subsection{Learning Rate Scaling} \label{ssec:lr_scale_analysis}

Former work in this space trained a 4096-d mLSTM with an initial learning rate of 5e-4, batches of 128, and a linear learning rate that decays to zero over one epoch of training \cite{Radford2017}. As expected and shown in Fig. \ref{tab:lr_scale}, keeping this same learning rate schedule as we increase batch size leads to worse accuracy, or a higher BPC\footnote{We train our models to convergence, not for one epoch, but results over one epoch are representative, and easier to compare.}.  

Recent work scaling image CNN models with SGD suggest that learning rates could be scaled linearly as batch size increases without a noticeable loss in model accuracy\cite{Smith2017a}. However, we found that our mLSTM model, optimized with Adam, diverged for large batches when we scaled up the 5e-4 initial learning rate with either a linear or a square root rule, as we increased the batch size.

We observed that for batch sizes of 2k-8k, the model converged reasonably well with an initial learning rate of 3e-3 decayed to zero over one epoch. Thus we kept 3e-3 as the initial learning rate for all other experiments. 
\begin{figure}[h!]
  \centering
    \begin{tabular}{l|r|r|r|r|r|r}
      \toprule % <-- Toprule here
      \textbf{Batch} & \textbf{Iters} & \textbf{Rule} & \textbf{LR} & \textbf{BPC} & \textbf{SST} & \textbf{IMDB}\\
      \midrule % <-- Midrule here
      \multirow{4}{*}{2048} & \multirow{4}{*}{72.6k} & linear & 8e-3 & 1.280 & 79.4 & 77.6 \\
      & & sqrt & 2e-3 & 1.117 & 90.2 & 91.9 \\
      & & - & 5e-4 & 1.130 & 89.1 & 90.8 \\
      & & - & 3e-3 & 1.110 & 89.0 & 92.1 \\
      \hline
      \multirow{4}{*}{4096} & \multirow{4}{*}{37.3k} & linear & 1.6e-2 & 1.275 & 78.3 & 77.6 \\
      & & sqrt & 2.8e-3 & 1.122 & 89.6 & 91.0 \\
      & & - & 5e-4 & 1.146 & 89.3 & 90.9 \\
      & & - & 3e-3 & 1.119 & 89.2 & 91.8 \\
      \hline
      \multirow{4}{*}{8192} & \multirow{4}{*}{18.6k} & linear & 3.2e-2 & 1.476 & 65.4 & 67.3 \\
      & & sqrt & 4e-3 & 1.133 & 89.7 & 90.8 \\
      & & - & 5e-4 & 1.175 & 87.3 & 89.6 \\
      & & - & 3e-3 & 1.132 & 89.5 & 91.4 \\
      \hline
      \multirow{4}{*}{16384} & \multirow{4}{*}{9.3k} & linear & 6.4e-2 & Div & - & - \\
      & & sqrt & 5.8e-3 & Div & - & - \\
      & & - & 5e-4 & 1.254 & 85.1 & 86.4 \\
      & & - & 3e-3 & 1.162 & 89.0 & 90.1 \\
      \hline
      \multirow{4}{*}{32768} & \multirow{4}{*}{4.6k} & linear & 1.3e-1 & Div & - & - \\
      & & sqrt & 8e-3 & Div & - & - \\
      & & - & 5e-4 & 1.380& 75.2 & 74.8 \\
      & & - & 3e-3 & 1.218 & 87.1 & 87.9 \\
      \bottomrule % <-- Bottomrule here
    \end{tabular}

    \caption{Evaluation results for various initial learning rates with a schedule decaying to zero over 1 epoch. Some initial rates are set by a linear or square root scaling rule based on a 5e-4 rate for a batch of 128. Div indicates that training diverged.}
    \label{tab:lr_scale}
\end{figure}

\subsection{Learning Rate Schedule} \label{ssec:lr_schedule}

When training to convergence, we used the same learning rate schedule for all batch sizes:
\begin{itemize}
    \item Set an initial learning rate of 3e-3.
    \item Linearly decay learning rate to zero over 100,000 iterations.
    \item Stop training at 3 epochs over the dataset, if fewer than 100,000 iterations.
\end{itemize}

This schedule, constant across all batch sizes, avoided the divergence observed in Fig. \ref{tab:lr_scale}, from scaling learning rate too much, but it also performed better than if we had instead kept the initial 5e-4 learning rate constant across batch sizes.

\begin{figure}[t!]
\captionsetup[subfigure]{justification=raggedright,singlelinecheck=false}
  \centering
  \begin{subfigure}[b]{\columnwidth}
  \caption{\label{long_run:graph}}
  \centering
\includegraphics[width=\columnwidth]{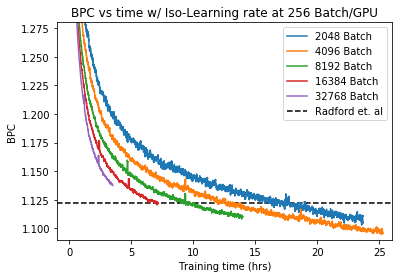}
  \end{subfigure}
  \begin{subfigure}[b]{\columnwidth}
   \caption{\label{long_run:table}}
   \centering
    \begin{tabular}{r|r|r|r|r|r|r|r}
      \toprule % <-- Toprule here
      \textbf{Batch} & \textbf{GPU} & \textbf{Iters} & \textbf{Ep} & \textbf{hrs} & \textbf{BPC} & \textbf{SST} & \textbf{IMDB}\\
      
      \midrule % <-- Midrule here
      2048 & 8 & 100k & 1.4 & 23.7 & 1.102 & 90.6 & 92.1 \\
      4096 & 16 & 100k & 2.7 & 25.3 & 1.090 & 90.6 & 92.7 \\
      8192 & 32 & 55k & 3.0 & 14.0 & 1.104 & 91.2 & 92.3 \\     
      16384 & 64 & 28k & 3.0 & 7.1 & 1.116 & 90.3 & 92.3 \\   
      32768 & 128 & 14k & 3.0 & 3.5 & 1.132 & 90.1 & 90.4 \\
      \bottomrule % <-- Bottomrule here
    \end{tabular}
  \end{subfigure}
  \caption{\subref{long_run:graph}) Training progress as a function of time for a 3e-3 initial learning rate decaying to zero over 100k iterations. Various batch sizes are trained with distributed data parallelism with a batch size of 256 per GPU. 
  \subref{long_run:table}) Comparison of model convergence, hardware used, time taken, and iterations and epochs (Ep) trained for a particular batch size. Batch sizes that have not reached 100k iterations after 3 epochs did not fully decay their learning rate and may benefit from more training.}
    \label{long_run}
\end{figure}

Using this learning rate schedule for the model with different batch sizes, Fig. \ref{long_run} shows that large batch training for this problem can converge to a similar evaluation BPC as smaller batch training given a good training schedule. However, adjusting the learning rate schedule is not as simple as modifying the learning rate according to batch size. In our experiments we found that controlling the steepness of decay was also required.

\section{Discussion}

We were able to converge our model in mixed precision, to a similar value as the FP32 baseline. This speeds up training, and substantially reduces our memory footprint, without a measurable change in accuracy, as shown in  Fig. \ref{tab:fp16_fp32}. We further speed up training by saturating up to 128 GPUs with distributed data parallelism, which we can do with a near-linear scaling factor Fig. \ref{tab:gpu_scale}.

However, as batch size increases from 128 in \cite{Radford2017} to 32k, the model needs more training steps to converge, and it does not converge to to quite as good a validation BPC as the low-batch model Fig. \ref{long_run}.

With longer training: 3 epochs of the Amazon Reviews dataset rather than 1, we do converge the 32k batch model close to the small batch model, doing so in a few hours instead of days or weeks. We also show that downstream task transfer to sentiment extraction is comparable when using converged large batch models (Fig. \ref{long_run:table}).

Smith et. al \cite{Smith2017a} suggest that to scale the learning rate without a loss in generalization quality, given a batch size $B$ and total amount of data $N$, $B$ must be sufficiently large so that $N\gg B$. It is possible that a batch size $\geq$ 32k and the amount of available Amazon data do not satisfy this requirement since the Amazon Reviews dataset is reduced to fewer than $5000$ iterations when we scale up to a 32k batch size. This observation opens up new research questions for future work.

\section{Future Work}

We have shown that distributed data parallelism scales for large RNN text models. However, we start to see diminishing returns on wall time convergence at very large batches, possibly because each epoch is reduced to a small number of training iterations. Now that we can train a language model on the 40 GB Amazon reviews dataset in hours, a next step could be to train on larger text datasets. Orders of magnitude larger text datasets could be constructed by collecting web pages, news articles, Reddit comments, and tweets, for example. 

In addition to larger text datasets, we could further improve Amazon Reviews BPC (and presumably accuracy on transfer tasks) with some of the following:
\begin{itemize}
   \item Training for more than 3 epochs.
   \item Data shuffling between epochs.
   \item Larger RNN models, with more layers and larger hidden states.
   \item Alternative language models, such as the Transformer network \cite{Transformer2017}.
   \item Hyper-parameter search for an ideal large batch learning rate schedule. 
\end{itemize}

As shown in Fig. \ref{long_run:table}, our best large batch training runs did not decay the learning rate to zero by the end of 3 epochs. As long as the initial learning rate is low enough not to cause training divergence, it may be possible to keep the learning rate high through several epochs of training. Recent language modeling work with the Transformer network has  shown that triangular learning rate warmup and non-linear learning rate decay (cosine annealing) can lead to a better learning rate schedule with the Adam optimizer\cite{Radford2018}. We showed that a simple learning rate schedule can work for large batch training, but further work on learning rate schedules will likely improve convergence.

Increasing the mLSTM size from 4096 to 8192-d reduces the per-GPU batch size by a factor of four. Using gradient checkpointing~\cite{checkpointing} would allow training larger models with larger batches without being constrained by memory capacity.

In order to get maximal text understanding from these larger models, we could modify the unsupervised task to include additional objectives, along with language modeling. Auxiliary tasks may include predicting a review's star rating, the title or topic of a piece of text, or any other freely available structural text label. Since the purpose of unsupervised training is to build a model with deep conceptual understanding of the text, auxiliary tasks that leverage metadata available with the text could provide additional understanding.

\section{Conclusion}

We set out to investigate large scale training for recurrent models in the Natural Language domain. With  mixed precision training we can successfully converge a model 4.2x faster with double the batch size compared to FP32 training. By leveraging distributed deep learning with NCCL2, NVLINK, and Infiniband interconnect, we achieve near linear scaling of 109x with 128 GPUs, as we grow the batch size proportionately to the number of available machines.

In addition to pushing wall time scalability by decreasing the time needed to converge a language model on the Amazon Reviews dataset, we analyze the convergence of models trained with large batches. We find that training with very large batches leads to somewhat worse generalization, requiring more data to converge to a similar validation BPC and transfer accuracy as small batch training. Learning rate schedule modifications are necessary to help with convergence. Without such techniques evaluation quality begins to decline as batch size increases, or the model fails to converge if the learning rate is scaled too high.

With further modification to the learning rate schedule and additional training it is possible to train models with large batches comparable to models trained with smaller batches. Our experiments lead to two insights:
\begin{itemize}
\item The relationship between batch size and learning regime is complex and learning rate scaling alone is not always enough to converge a model.
\item Even with the largest public text corpus available, it may not be feasible to satisfy the $B \ll N$ batch size requirement needed to effectively train with the largest batches that modern hardware allows.
\end{itemize}

We look forward to more work investigating large scale language model training and using it in transferred tasks to solve difficult natural language problems. 

\bibliography{HPML}
\bibliographystyle{HPML}

\end{document}